\documentclass[10pt,sigconf]{acmart}

\AtBeginDocument{%
  \providecommand\BibTeX{{%
    \normalfont B\kern-0.5em{\scshape i\kern-0.25em b}\kern-0.8em\TeX}}}

\copyrightyear{2020}
\acmYear{2020}
\setcopyright{acmlicensed}
\acmConference[SIGMOD'20]{Proceedings of the 2020 ACM SIGMOD International Conference on Management of Data}{June 14--19, 2020}{Portland, OR, USA}
\acmBooktitle{Proceedings of the 2020 ACM SIGMOD International Conference on Management of Data (SIGMOD'20), June 14--19, 2020, Portland, OR, USA}
\acmPrice{15.00}
\acmDOI{10.1145/3318464.3386145}
\acmISBN{978-1-4503-6735-6/20/06}

\usepackage{booktabs} 
\usepackage{balance}

\usepackage[linesnumbered,ruled,vlined]{algorithm2e} 

\SetAlFnt{\small}
\SetAlCapFnt{\small}
\SetAlCapNameFnt{\small}
\SetAlCapHSkip{0pt}
\IncMargin{-\parindent}

\usepackage{tabularx}
\usepackage{multirow}
\usepackage{subfigure}
\usepackage{CJKutf8}
\usepackage[utf8]{inputenc}
\usepackage[T1]{fontenc}
\usepackage{subfigure}
\usepackage{amsmath, amssymb}
\usepackage{algorithmic}
\usepackage{tabularx}
\usepackage{color}
\SetKwInput{kwClusterStep}{Query-Doc Clustering Step}
\SetKwInput{kwExtractStep}{Attention Phrase Mining Step}
\SetKwInput{kwNormalizeStep}{Attention Phrase Normalization Step}
\SetKwInput{kwDerivingStep}{Attention Phrase Derivation Step}
\definecolor{pink}{rgb}{0.858, 0.188, 0.478}
\definecolor{orange}{rgb}{1.0, 0.35, 0.0}

\begin{document}

\fancyhead{}
\title{GIANT: Scalable Creation of a Web-scale Ontology}

\author{Bang Liu$^{1*}$, Weidong Guo$^{2*}$, Di Niu$^1$, Jinwen Luo$^2$}
\author{Chaoyue Wang$^2$, Zhen Wen$^2$, Yu Xu$^2$}
\thanks{$^*$Equal contribution. Correspondence to: weidongguo@tencent.com}
\affiliation{$^1$University of Alberta, Edmonton, AB, Canada}
\affiliation{$^2$Platform and Content Group, Tencent, Shenzhen, China}

\begin{abstract}
Understanding what online users may pay attention to is key to content recommendation and search services. These services will benefit from a highly structured and web-scale ontology of entities, concepts, events, topics and categories. While existing knowledge bases and taxonomies embody a large volume of entities and categories, we argue that they fail to discover properly grained concepts, events and topics in the language style of online population. Neither is a logically structured ontology maintained among these notions. In this paper, we present \textit{GIANT}, a mechanism to construct a user-centered, web-scale, structured ontology, containing a large number of natural language phrases conforming to user attentions at various granularities, mined from a vast volume of web documents and search click graphs. Various types of edges are also constructed to maintain a hierarchy in the ontology. We present our graph-neural-network-based techniques used in GIANT, and evaluate the proposed methods as compared to a variety of baselines. GIANT has produced the Attention Ontology, which has been deployed in various Tencent applications involving over a billion users. Online A/B testing performed on Tencent QQ Browser shows that Attention Ontology can significantly improve click-through rates in news recommendation.
\end{abstract} 

%
%

\ccsdesc[500]{Information systems~Data mining}
\ccsdesc[500]{Information systems~Document representation}
\ccsdesc[500]{Computing methodologies~Information extraction}

\keywords{Ontology Creation, Concept Mining, Event Mining, User Interest Modeling, Document Understanding}

\maketitle

\section{Introduction}
\label{sec:intro}

In a fast-paced society, most people carefully choose what they pay attention to in their overstimulated daily lives.
With today's information explosion, it has become increasingly challenging to increase the attention span of online users. 
A variety of recommendation services \cite{konstan2008introduction,adomavicius2005toward,koutrika2018modern,bobadilla2013recommender,adomavicius2011context} have been designed and built to recommend relevant information to online users based on their search queries or viewing histories.
Despite a variety of innovations and efforts that have been made, these systems still suffer from two long-standing problems, inaccurate recommendation and monotonous recommendation.

Inaccurate recommendation is mainly attributed to the fact that most content recommender, e.g., news recommender, are based on keyword matching.
For example, if a user reads an article on ``Theresa May's resignation speech'', current news recommenders will try to further retrieve articles that contain the keyword ``Theresa May'', although the user is most likely not interested in the person ``Theresa May'', but instead is actually concerned with ``Brexit negotiation'' as a topic, to which the event ``Theresa May's resignation speech'' belongs.
Therefore, keywords that appear in an article may not always be able to characterize a user's interest. Frequently, a higher and proper level of abstraction of verbatim keywords is helpful to recommendation, e.g., knowing that Honda Civic is an ``economy car'' or ``fuel-efficient car'' is more helpful than knowing it is a sedan. 

Monotonous recommendation is the scenario where users are recommended with articles that always describe the same entities or events repeatedly. This phenomenon is rooted in the incapability of existing systems to extrapolate beyond the verbatim meaning of a viewed article. Taking the above example again, instead of pushing similar news articles about ``Theresa May's resignation speech'' to the same user, a way to break out from redundant monotonous recommendation is to recognize that ``Theresa May's resignation speech'' is an event in the bigger topic ``Brexit negotiations'' and find other preceding or follow-up events in the same topic.

To overcome the above-mentioned deficiencies, what is required is an ontology of ``user-centered'' terms discovered at the web scale that can abstract keywords into concepts and events into topics in the vast open domain, such that user interests can be represented and characterized at different granularities, while maintaining a structured hierarchy among these terms to facilitate interest inference.
However, constructing an ontology of user interests, including entities, concepts, events, topics and categories, from the open web is a very challenging task.
Most existing taxonomy or knowledge bases, such as Probase \cite{wu2012probase}, DBPedia \cite{lehmann2015dbpedia}, CN-DBPedia \cite{xu2017cn}, YAGO \cite {suchanek2007yago}, extract concepts and entities from Wikipedia and web pages based on Hearst patterns, e.g., by mining phrases around ``such as'', ``especially'', ``including'' etc.
However, concepts extracted this way are clearly limited. 
Moreover, web pages like Wikipedia are written in an author's perspective and are not the best at tracking user-centered Internet phrases like ``top-5 restaurants for families''. 

To mine events, most existing methods \cite{aone2000rees,miwa2010event,mcclosky2011event,yang2016joint} rely on the \emph{ACE (Automatic Content Extraction) definition} \cite{doddington2004automatic,grishman2005nyu} and perform event extraction within specific domains via supervised learning by following predefined patterns of triggers and arguments. This is not applicable to vastly diverse types of events in the open domain.
There are also works attempting to extract events from social networks such as Twitter \cite{watanabe2011jasmine,ritter2012open,atefeh2015survey,cordeiro2016online}. But they represent events by clusters of keywords or phrases, without identifying a clear hierarchy or  ontology among phrases.

In this paper, we propose \textbf{GIANT}, 
a web-based, structured ontology construction mechanism that can automatically discover critical natural language phrases or terms, which we call user \emph{attentions}, that characterize user interests at different granularities, by mining unstructured web documents and search query logs at the web scale. In the meantime, GIANT also aims to form a hierarchy of the discovered user attention phrases to facilitate inference and extrapolation.
GIANT produces and maintains a web-scale ontology, named the \emph{Attention Ontology} (AO), which consists of around 2 million nodes of five types, including categories, concepts, entities, topics, and events, and is still growing.
In addition to categories (e.g., 
cars, technology) and entities (e.g., 
iPhone XS, Honda Civic) that can be commonly found in existing taxonomies or knowledge graphs,  Attention Ontology also contains abstractive terms at various semantic granularities, including newly discovered \emph{concepts}, \emph{events} and \emph{topics}, all in user language or popular online terms. 
By tagging documents with these abstractive terms that the online population would actually pay attention to, Attention Ontology proves to be effective in improving content recommendation in the open domain.
For example, with the ontology, the system can extrapolate onto the concept ``economy cars'' if a user has browsed an article about ``Honda Civic'', even if the exact wording of ``economy cars'' does not appear in that article.
The system can also infer a user's interest in all events related to the topic ``Brexit Negotiation'' if he or she viewed an article about ``Theresa May's resignation''.

GIANT constructs the user-centered Attention Ontology by mining the \emph{click graph}, a large bipartite graph formed by search queries and their corresponding clicked documents.
GIANT relies on the linkage from queries to the clicked documents to discover terms that may represent user attentions.
For example, if a query ``top-5 family restaurants in Vancouver'' frequently leads to the clicks on certain restaurants, we can recognize ``top-5 family restaurants in Vancouver'' as a concept and the clicked restaurants are entities within this concept.
Compared to existing knowledge graphs constructed from Wikipedia, which contain general factual knowledge, queries from users can reflect hot concepts, events or topics that are of user interests at the present time.

To automatically extract concepts, events or topics from the click graph, we propose  GCTSP-Net (Graph Convolution-Traveling Salesman Problem Network), a multi-task model which can mine different types of phrases at scale.
Our model is able to extract heterogeneous phrases in a unified manner.
The extracted phrases become the nodes in the Attention Ontology, which can properly summarize  true user interests and can also be utilized to characterize the main theme of queries and documents.
Furthermore, to maintain a hierarchy within the ontology, we propose various machine learning and ad-hoc approaches to identify the edges between nodes in the Attention Ontology and tag each edge with one of the three types of relationships: \textit{isA}, \textit{involve}, and \textit{correlate}.
The constructed edges can benefit a variety of applications, e.g., concept-tagging for documents at an abstractive level, query conceptualization, event-series tracking, etc.






\begin{figure*}
\includegraphics[width=\textwidth]{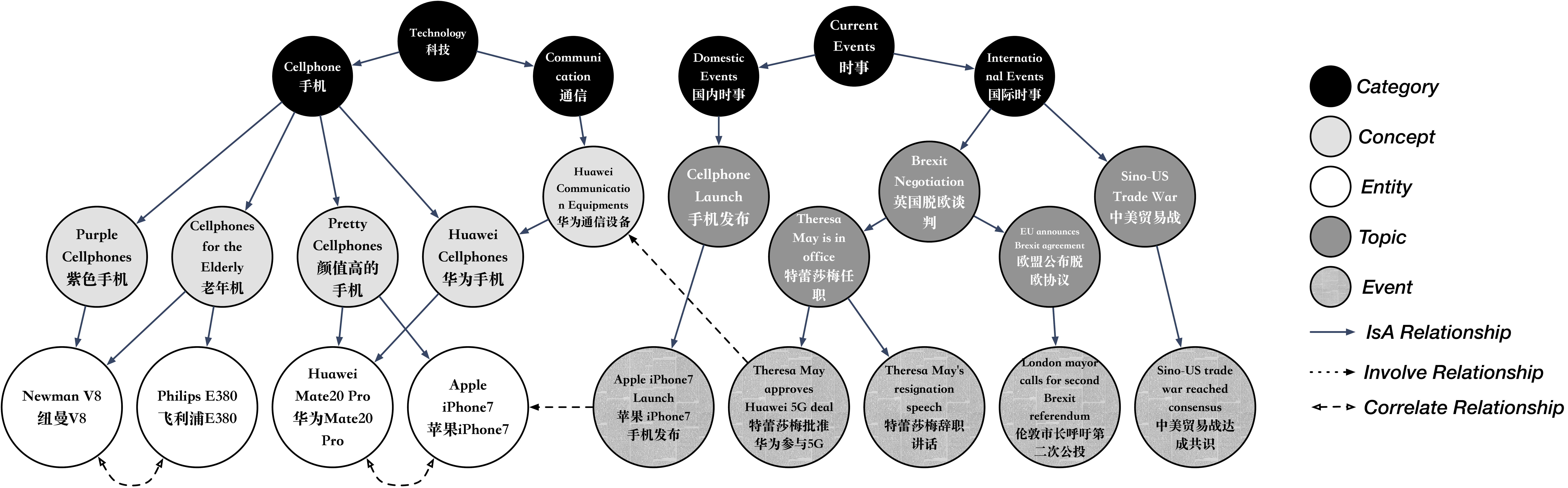}
\caption{An example to illustrate our \textit{Attention Ontology} (AO) for user-centered text understanding.}
\Description[Attention Ontology]{An example to illustrate our \textit{Attention Ontology} (AO) for user-centered text understanding.}
\label{fig:ag}
\vspace{-4mm}
\end{figure*}

We have performed extensive evaluation of GIANT.
For all the learning components in GIANT, we introduce efficient strategies to quickly build the training data necessary to perform phrase mining and relation identification, with minimum manual labelling efforts. For phrase mining, we combine a bootstrapping strategy based on pattern-phrase duality \cite{brin1998extracting,liu2019concept} with text alignment based on query-title conformity. 
For relationship identification, we utilize the co-occurrence of different phrases in queries, documents, and consecutive queries to extract phrase pairs labeled with target relationships. These labeled examples automatically mined from the click graph are then used to train the proposed machine learning models in different sub-tasks.

The assessment of the constructed ontology shows that our system can extract a plethora of high-quality phrases, with a large amount of correctly identified relationships between these phrases.
We compare the proposed GNN-based GCTSP-Net with a variety of baseline approaches to evaluate its performance and superiority on multiple tasks.
The experimental results show that our approach can extract heterogeneous phrases more accurately from the click graph as compared to existing methods.


It is worth noting that GIANT has been deployed in multiple real-world applications, including Tencent QQ browser, Mobile QQ and WeChat, involving more than 1 billion active users around the globe, and currently serves as the core taxonomy construction system in these commercial applications to discover long-term and trending user attentions. We report the online A/B testing results of introducing Attention Ontology into Tencent QQ Browser mobile app, which is a news feed app that displays news articles as a user scrolls down on her phone. The results suggest that the generated Attention Ontology can significantly improve the click-through rate in news feeds recommendation.


\section{The Attention Ontology}
\label{sec:overview}


In the proposed Attention Ontology, each node is represented by a phrase or a collection of phrases of the same meaning mined from Internet.
We use the term ``attention'' as a general name for \emph{entities}, \emph{concepts}, \emph{events}, \emph{topics}, and \emph{categories}, which represent five types of information that can capture an online user's attention at different semantic granularities. 
Such attention phrases can be utilized to conceptualize user interests and depict document coverages.
For instance, if a user wants to buy a family vehicle for road trips, he/she may input such a query ``vehicles choices for family road trip''.
From this query, we could extract the concept, ``family road trip vehicles'', and tag it to matching entities such as ``Honda Odyssey Minivan'' or ``Ford Edge SUV''. We could then recommend articles related to these Honda and Ford vehicles, even if they do not contain the exact wording of ``family road trip''. 
In essence, the Attention Ontology enables us to achieve a user-centered understanding of web documents and queries, which improves the performance of search engines and recommender systems. 

Figure~\ref{fig:ag} shows an illustration of the Attention Ontology, which is in the form of a Directed Acyclic Graph (DAG). Specifically,  there are five types of nodes:





\begin{itemize}
\item \textbf{Category}: a category node defines a broad field that encapsulates many related topics or concepts. For example, technology, current events, entertainment, sports, finance and so forth. In our system, we pre-define a 3-level categories hierarchy, which consists of 1,206 different categories.

\item \textbf{Concept}: a concept is a group of entities that share some common attributes \cite{liu2019concept,wu2012probase}, such as superheroes, MOBA games, fuel-efficient cars and so on. In contrast with coarse-grained categories and fine-grained entities, concepts can help better characterize users' interests at a suitable semantic granularity.

\item \textbf{Entity}: an entity is a specific instance belonging to one or multiple concepts. For example, Iron Man is an entity belonging to the concepts ``superheroes'' and ``Marvel superheroes''.

\item \textbf{Event}: an event is a real-world incident that involves a group of specific persons, organizations, or entities. It is also tagged with a certain time/location of occurrence \cite{liu2017growing}.
In our work, we represent an event with four types of attributes: \textit{entities} (indicating who or what are involved in the event), \textit{triggers} (indicating what kind/type of event it is), \textit{time} (indicating when the event takes place), and \textit{location} (indicating where the event takes place).

\item \textbf{Topic}:
a topic represents a collection of events sharing some common attributes. For example, both ``Samsung Galaxy Note 7 Explosion in China'' and ``iPhone 6 Explosion in California'' events belong to the topic ``Cellphone Explosion''. 
Similarly, events such as ``Theresa May is in Office'', ``Theresa May's Resignation Speech'' can be covered by the topic ``Brexit Negotiation''.
\end{itemize}
Furthermore, we define three types of edges, i.e., relationships, between nodes:
\begin{itemize}
	\item \textbf{isA relationship}, indicating that the destination node is an instance of the source node. For example, the entity ``Huawei Mate20 Pro" \emph{isAn} instance of ``Huawei Cellphones".
	\item \textbf{involve relationship}, indicating that the destination node is involved in an event/topic described by the source node.
	\item \textbf{correlate relationship}, indicating two nodes are highly correlated with each other.
\end{itemize}

The edges in the Attention Ontology reveal the types of relationships and the degrees of relatedness between nodes. A plethora of edges enables the inference of more hidden interests of a user beyond the content he/she has browsed by moving along the edges on the Attention Ontology and recommending other related nodes at a coarser or finer granularity based on the nodes the user has visited.
Furthermore, by analyzing edges between event nodes, we could also keep track of a \emph{developing story}, which usually consists of a series of events, and keep interested users updated.

\section{Ontology Construction}
\label{sec:attention}

GIANT is a mechanism to discover phrases that users pay attention to from the web as well as to build a structured hierarchy out of them. 
In this section, we present our detailed techniques to construct the Attention Ontology based on neural networks and other ad-hoc methods.
The entire process consists of two phases: i) user attention phrases mining, and ii)  attention phrases linking.
In the first phase, we define and extract user attention phrases in different semantic granularities from a large-scale search click graph.
In the second phase, we link different extracted nodes and identify their relationships to construct a structured ontology.


\begin{figure}
\includegraphics[width=0.5\textwidth]{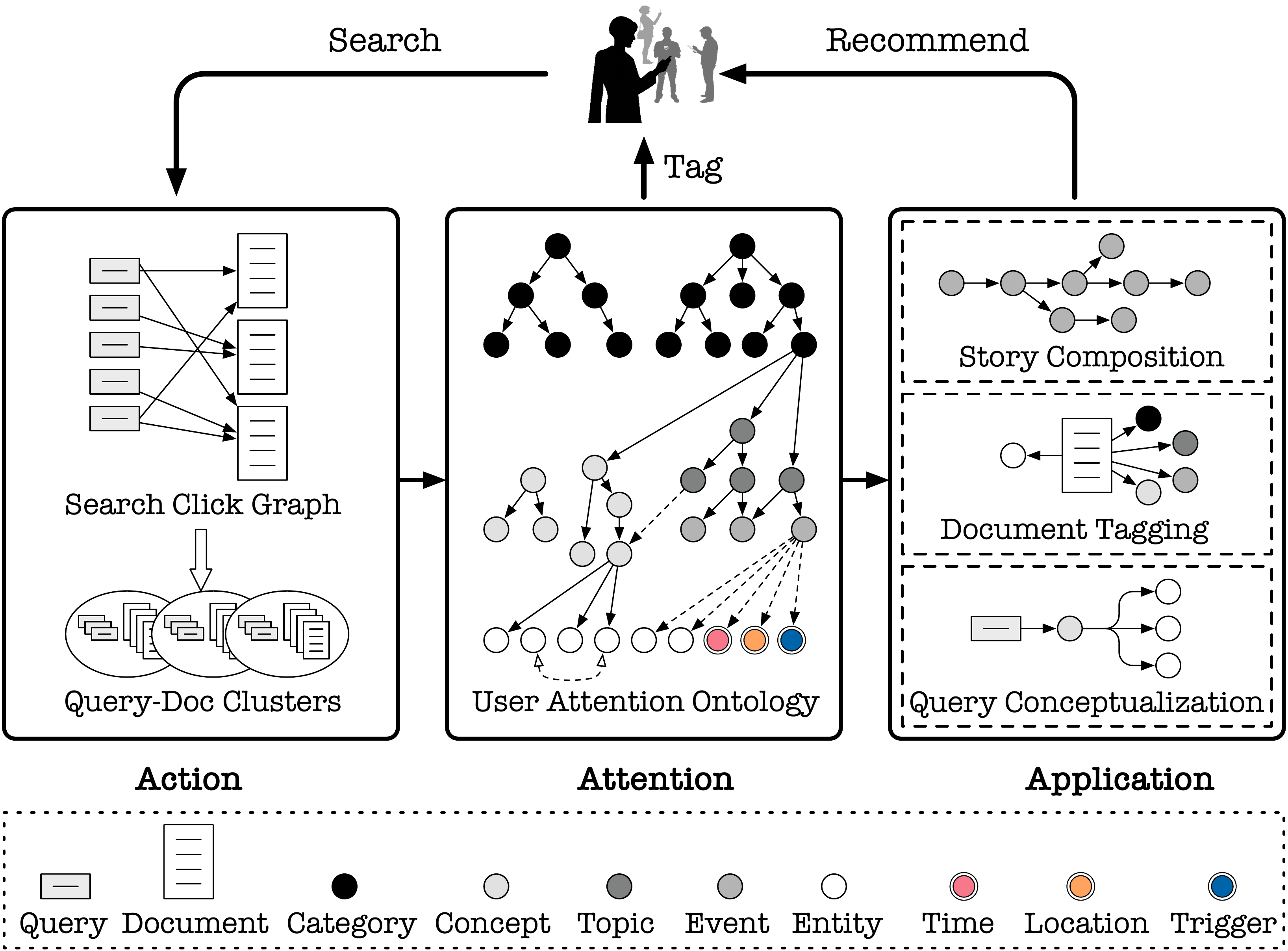} 
\caption{Overview of our framework for constructing the Attention Ontology and performing different tasks.}
\Description[Framework Overview]{Overview of our \textit{A3} framework for constructing attention graph and performing different tasks.}
\label{fig:A3}
\vspace{-4mm}
\end{figure}

Figure~\ref{fig:A3} shows the overall framework of GIANT, which constructs the Attention Ontology based on user search and click logs.
The framework consists of three major components: action, attention, and application.
In the action component, when users perform different actions (e.g., search, click, etc.), the user queries and clicked documents form a bipartite graph \cite{wiki:bipartiteG}, commonly known as a search click graph. Based on it, we can collect highly correlated queries and documents by aggregating documents that correspond to a query or vice versa, into query-doc clusters.
In the attention component, we can extract different types of nodes (e.g., concepts, events, topics, entities, etc.) from the query-doc clusters, as well as learn the relationships between different nodes to form the Attention Ontology.
In the application component, we can apply the Attention Ontology to different applications such as query conceptualization and document tagging. On the other hand, we can also integrate different nodes to user profiles to characterize the interest of different users based on his/her historical viewing behavior. In this manner, we can characterize both users and documents based on the Attention Ontology, which enables us to better understand and recommend documents from users' perspective.

\subsection{Mining User Attentions}
\label{sec:mining}


We propose a novel algorithm to extract various types of attention phrases (e.g., concepts, topics or events), which represent user attentions or interests, from a collection of queries and document titles.

\textbf{Problem Definition}.
Suppose a bipartite search click graph $G_{sc} = (Q, D, E)$ records the click-through information over a set of queries $Q=\{q_1, q_2, \cdots, q_{|Q|}\}$ and documents $D=\{d_1, d_2, \cdots, d_{|D|}\}$. We use $|*|$ to denote the length of $*$. $E$ is a set of edges linking queries and documents.
Our objective is to extract a set of phrases $P=\{p_1, p_2, \cdots, p_{|P|}\}$ from $Q$ and $D$ to represent user interests.
Specifically, suppose $p$ consists of a sequence of words $\{w_{p1}, w_{p2}, \cdots, w_{p|p|}\}$. In our work, each phrase $p$ is extracted from a subset of queries and the titles of correlated documents, and each word $w_p \in p$ is contained by at least one query or title.

\begin{algorithm}[h]
\KwIn{a sequence of queries $Q= \{q_1, q_2, \cdots, q_{|Q|}\}$, search click graph $G_{sc}$.}
\KwOut{Attention phrases $P = \{p_1, p_2, \cdots, p_{|P|}\}$.}

\begin{algorithmic}[1]
\STATE calculating the transport probabilities between connected query-doc pairs according to Equation \eqref{eq:qdcluster1} and \eqref{eq:qdcluster}\;
\FOR{each $q \in Q$}
    \STATE run random walk to get ordered related queries $Q_q$ and documents $D_q$\;
\ENDFOR

\STATE $P = \{\}$\;
\FOR{each input cluster $(Q_q, D_q)$}
    \STATE get document titles $T_q$ from $D_q$\; 
    \STATE create Query-Title Interaction Graph $G_{qt}(Q_q, T_q)$\;
    \STATE classify the nodes in $G_{qt}(Q_q, T_q)$ by R-GCN to learn which nodes belong to the output phrase\;
    \STATE sort the extracted nodes by ATSP-decoding and concatenate them into an attention phrase $p^a_q$\;
    \STATE perform attention normalization to merge $p^a_q$ with its similar phrase in $P$ into a sublist\;
    \STATE if $p^a_q$ is not similar to any existing phrase, append $p^a_q$ to $P$\;
\ENDFOR

\STATE create a node in the Attention Ontology for each phrase or sublist of similar phrases.
\end{algorithmic}

\caption{Mining Attention Nodes}
\label{alg:attention-mine}
\end{algorithm}

Algorithm~\ref{alg:attention-mine} presents the pipeline of our system to extract attention phrases based on a bipartite search click graph. In what follows, we introduce each step in detail.

\textbf{Query-Doc Clustering}.
Suppose $c(q_i, d_j)$ represents how many times query $q_i$ is linked to document $d_j$ in a search click graph $G_{sc}$ constructed from user search click logs within a period. For each query-doc pair $<q_i, d_j>$, suppose $N(q_i)$ denotes the set of documents connected with $q_i$, and $N(d_j)$ denotes the set of queries connected with $d_j$. Then we define the transport probabilities between $q_i$ and $d_j$ as:
\begin{align}
\mathbb{P}(d_j | q_i) &= \frac{c(q_i, d_j)}{\sum_{d_k \in N(q_i)} c(q_i, d_k)}, \label{eq:qdcluster1}\\
\mathbb{P}(q_i | d_j) &= \frac{c(q_i, d_j)}{\sum_{q_k \in N(d_j)} c(q_k, d_j)}.
\label{eq:qdcluster}
\end{align}
From query $q$, we perform random walk \cite{spitzer2013principles} according to transport probabilities calculated above and compute the weights of visited queries and documents.
For each visited query or document, we keep it if its visiting probability is above a threshold $\delta_v$ and the number of non-stop words in $q$ is more than a half. In this way, we can derive a cluster of correlated queries $Q_q$ and documents $D_q$.

\begin{figure}
\includegraphics[width=0.46\textwidth]{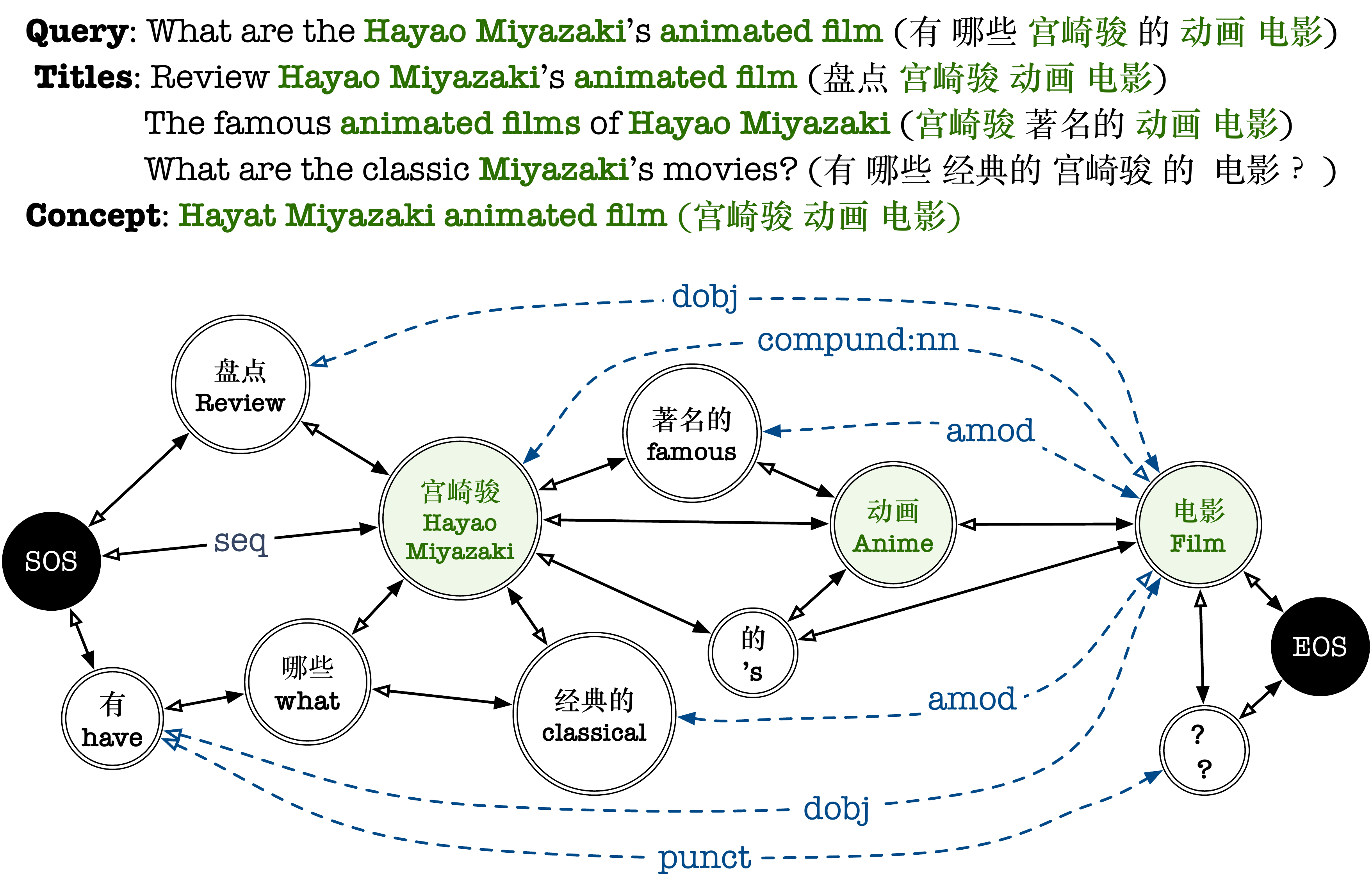} 
\caption{An example to show the construction of query-title interaction graph for attention mining.}
\Description[Node Extraction]{An example to show the construction of query-title interaction graph for attention mining.}
\label{fig:node-extract}
\vspace{-4mm}
\end{figure}

\textbf{Query-Title Interaction Graph Construction}.
Given a set of queries $Q_q$ and $D_q$, the next step is to extract a representative phrase that captures the underlying user attentions or interests.
Figure~\ref{fig:node-extract} shows a query-doc cluster and the concept phrase extracted from it.
We can extract `Hayao Miyazaki animated film (\begin{CJK}{UTF8}{gkai}宫崎骏|动画|电影\end{CJK})'' from the input query-title cluster.
An attention phrase features multiple characteristics.
First, the tokens in it may show up multiple times in the queries and document titles.
Second, the phrase tokens are not necessarily consecutively or fully contained by a single query or title.
For example, in Figure~\ref{fig:node-extract}, additional tokens such as ``famous (\begin{CJK}{UTF8}{gkai}著名的\end{CJK})'' will be inserted into the phrase tokens, making them not a consecutive span.
Third, the phrase tokens are syntactically dependent even if they are not consecutively adjacent in a query or title.
For example, ``Hayao Miyazaki (\begin{CJK}{UTF8}{gkai}宫崎骏\end{CJK})'' and ``film (\begin{CJK}{UTF8}{gkai}电影\end{CJK})'' forms a compound noun name.
Finally, the order of phrase tokens may change in different text.
In Figure~\ref{fig:node-extract}, the tokens in the concept phrase ``Hayao Miyazaki animated film (\begin{CJK}{UTF8}{gkai}宫崎骏|动画|电影\end{CJK})'' are fully contained by the query and two titles, while the order of the tokens varies in different queries or titles.
Other types of attention phrases such as events and topics will feature similar characteristics.

To fully exploit the characteristics of attention phrases, we propose Query-Title Interaction Graph (QTIG), a novel graph representation of queries and titles to reveal the correlations between their tokens.
Based on it, we further propose GCTSP-Net, a model that takes a query-title interaction graph as input, performs node classification with graph convolution, and finally generates a phrase by Asymmetric Traveling Salesman Decoding (ATSP-decoding).

Figure~\ref{fig:node-extract} shows an example of query-title interaction graph constructed from a set of queries and titles.
Denote a QTIG constructed from queries $Q_q$ and the titles $T_q$ of documents $D_q$ as $G_{qt}(Q_q, T_q)$. 
The queries and documents are sorted by the weights calculated during the random walk.
In $G_{qt}(Q_q, T_q)$, each node is a unique token belonging to $Q_q$ or $T_q$.
The same token present in different input text will be merged into one node.
For each pair of nodes, if they are adjacent tokens in any query or title, they will be connected by a bi-directional ``{\tt seq}'' edge, indicating their order in the input.
In Figure~\ref{fig:node-extract}, the inverse direction of a ``seq'' edge points to the preceding words, which is indicated by a hollow triangle pointer.
If the pair of nodes are not adjacent to each other, but there exists syntactical dependency between them, they will be connected by a bi-directional dashed edge which indicates the type of syntactical dependency relationship and the direction of it, while the inverse direction is also indicated by a hollow triangle pointer.

\begin{algorithm}[h]
\KwIn{a sequence of queries $Q= \{q_1, q_2, \cdots, q_{|Q|}\}$, document titles $T = \{t_1, t_2, \cdots, t_{|T|}\}$.}
\KwOut{a Query-Title Interaction Graph $G_{qt}(Q, T)$.}

\begin{algorithmic}[1]
\STATE Create node set $V = \{ sos, eos \}$, edge set $E = \{\}$\;
\FOR{each input text passage $x \in Q$ or $x \in T$}
    \STATE append ``{\tt sos}'' and ``{\tt eos}'' as the first and the last token of $x$\;  
    \STATE construct a new node for each token in $x$\;
    \STATE construct a bi-directional ``{\tt seq}'' edge for each pair of adjacent tokens\;
    \STATE append each constructed node and edge into $V$ or $E$ only if the node is not contained by $V$, or no edge with the same source and target tokens exists in $E$\;
\ENDFOR
\FOR{each input text passage $x \in Q$ or $x \in T$}
    \STATE perform syntactic parsing over $x$\;
    \STATE construct a bi-directional edge for each dependency relationship\;
    \STATE append each constructed edge into $E$  if no edge with the same source and target tokens exists in $E$\;
\ENDFOR
\STATE construct graph $G_{qt}(Q, T)$ from node set $V$ and edge set $E$.
\end{algorithmic}

\caption{Constructing the Query-Title Interaction Graph}
\label{alg:qtig}
\end{algorithm}

Algorithm~\ref{alg:qtig} shows the process of constructing a query-title interaction graph.
We construct the nodes and edges by reading the inputs in $Q_q$ and $T_q$ in order.
When we construct the edges between two nodes, as two nodes may have multiple adjacent relationships or syntactical relationships in different inputs, we only keep the first edge  constructed.
In this way, each pair of related nodes will only be connected by a bi-directional sequential edge or a syntactical edge.
The idea is that we prefer the ``{\tt seq}'' relationship as it shows a stronger connection than any syntactical dependency, and we prefer the syntactical relationships contained in the higher-weighted input text instead of the relationships in lower-weighted inputs.
Compared with including all possible edges in a query-title interaction graph, empirical evidence suggests that our graph construction approach gives better performance for phrase mining.

After constructing a graph $G_{qt}(Q_q, T_q)$ from a query-title cluster, we extract a phrase $p$ by our GCTSP-Net, which contains a classifier to predict whether a node belong to $p$, and an asymmetric traveling salesman decoder to order the predicted positive nodes.

\textbf{Node Classification with R-GCN}.
In the GCTSP-Net, we apply Relational Graph Convolutional Networks (R-GCN) \cite{kipf2016semi,gilmer2017neural,schlichtkrull2017modeling} to our constructed QTIG to perform node classification.

Denote a directed and labeled multi-graph as $G = (V, E, R)$ with labeled edges $e_{vw} = (v, r, w) \in E$, where $v, w \in V$ are connected nodes, and $r \in R$ is a relation type.
A class of graph convolutional networks can be understood as a message-passing framework \cite{gilmer2017neural}, where the hidden states $h^l_v$ of each node $v \in G$ at layer $l$ are updated based on messages $m^{l+1}_v$ according to:
\begin{align}
m^{l+1}_{v} &= \sum_{w\in N(v)} M_l(h^l_v, h^l_w, e_{vw}), \\
h^{l+1}_v &= U_l(h^l_v, m^{l+1}_v),
\end{align}
where $N(v)$ denotes the neighbors of $v$ in graph $G$, $M_l$ is the message function at layer $l$, and $U_l$ is the vertex update function at layer $l$.

Specifically, the message passing function of Relational Graph Convolutional Networks is defined as:
\begin{align}
h_v^{l+1} = \sigma\Bigg(\sum_{r \in R} \sum_{w\in N^r(v)} \frac{1}{c_{vw}} W^l_r h^l_w + W^l_0 h^l_v \Bigg),
\end{align}
where $\sigma(\cdot)$ is an element-wise activation function such as ReLU$(\cdot) = \text{max}(0, \cdot)$. $N^r(v)$ is the set of neighbors under relation $r\in R$. $c_{vw}$ is a problem-specific normalization constant that can be learned or pre-defined (e.g., $c_{vw} = |N^r(v)|$). $W^l_r$ and $W^l_0$ are learned weight matrices.

We can see that an R-GCN accumulates transformed feature vectors of neighboring nodes through a normalized sum. Besides, it learns relation-specific transformation matrices to take the type and direction of each edge into account.
In addition, it adds a single self-connection of a special relation type to each node to ensure that the representation of a node at layer $l + 1$ can also be informed by its representation at layer $l$.

To avoid the rapid growth of the number of parameters when increasing the number of relations $|R|$, R-GCN exploits basis decomposition and block-diagonal decomposition to regularize the weights of each layer.
For basis decomposition, each weight matrix $W^l_r \in \mathbb{R}^{d^{l+1}\times d^l}$ is decomposed as:
\begin{align}
W^l_r = \sum_{b=1}^{B} a^l_{rb} V^l_b,
\end{align}
where $V^l_b \in \mathbb{R}^{d^{l+1}\times d^l}$ are base weight matrices.
In this way, only the coefficients $a^l_{rb}$ depend on $r$. For block-diagonal decomposition, $W^l_r$ is defined through the direct sum over a set of low-dimensional matrices:
\begin{align}
W^l_r = \mathop{\bigoplus}_{b=1}^{B}Q^l_{br},
\end{align}
where $W^l_r = \text{diag}(Q^l_{1r}, Q^l_{2r}, \cdots, Q^l_{br})$ is a block-diagonal matrix with $Q^l_{br} \in \mathbb{R}^{(d^{l+1}/B) \times (d^{l}/B)}$.
The basis function decomposition introduces weight sharing between different relation types, while the block decomposition applies sparsity constraint on the weight matrices.

In our model, we apply R-GCN with basis decomposition to query-title interaction graphs to perform node classification.
For each node in the graph, we represent it by a feature vector consisting of the embeddings of the token's named entity recognition (NER) tag, part-of-speech (POS) tag, whether it is a stop word, number of characters in the token, as well as the sequential id that indicates the order each node be added to the graph.
Using the concatenation of these embeddings as the initial node vectors, we pass the graph to a multi-layer R-GCN with a $\text{softmax}(\cdot)$ activation (per node) on the output of the last layer.
We label the nodes belonging to the target phrase $p$ as $1$ and other nodes as $0$, and train the model by minimizing the binary cross-entropy loss on all nodes.

\textbf{Node Ordering with ATSP Decoding}.
After classified a set of tokens $V_p$ as target phrase nodes, the next step is to sort the nodes to get the final output.
In our GCTSP-Net, we propose to model the problem as an asymmetric traveling salesman problem (ATSP), where the objective is to find the shortest route that starts from the ``{\tt sos}'' node, visits each predicted positive nodes, and returns to the  ``{\tt eos}'' node.
This approach is named as ATSP-decoding in our work.

We perform ATSP-decoding with a variant of the constructed query-title interaction graph.
First, we remove all the syntactical dependency edges. Second, instead of connecting adjacent tokens by a bi-directional ``{\tt seq}'' edge, we change it into unidirectional to indicate the order in input sequences. Third, we connect ``{\tt sos}''  with the first predicted positive token in each input text, as well as connect the last predicted positive token in each input with the ``{\tt eos}'' node. In this way, we remove the influence of prefixing and suffixing tokens in the inputs when finding the shortest path. Finally, the distance between a pair of predicted nodes is defined as the length of the shortest path in the modified query-title interaction graph. In this way, we can solve the problem with Lin-Kernighan Traveling Salesman Heuristic \cite{helsgaun2000effective} to get a route of the predicted nodes and output $p$.

We shall note that ATSP-decoding will produce a phrase that contains only unique tokens.
In our work, we observe that only less than $1\%$ attention phrases contain duplicated tokens, while most of the duplicated tokens are punctuations.
Even if we need to produce duplicate tokens when applying our model to other tasks, we just need to design task-specific heuristics to recognize the potential tokens (such as count their frequency in each query and title), and construct multiple nodes for it in the query-title interaction graph.

\textbf{Attention Phrase Normalization}.
The same user attention may be expressed by slightly different phrases. After extracting a phrase using GCTSP-Net, we merge highly similar phrases into one node in out Attention Ontology.
Specifically, we examine whether a new phrase $p_{n}$ is similar to an existing phrase $p_e$ by two criteria: i) the non-stop words in $p_{n}$ shall be similar (same or synonyms) with that in $p_e$, and ii) the TF-IDF similarity between their context-enriched representations shall be above a threshold $\delta_m$. The context-enriched representation of a phrase is obtained by using itself as a query and concatenating the top $5$ clicked titles.

\textbf{Training Dataset Construction}.
To reduce human effort and accelerate the labeling process of training dataset creation, we design effective unsupervised strategies to extract candidate phrases from queries and titles, and provide the extracted candidates together with query-title clusters to workers as assistance.
For concepts, we combine bootstrapping with query-title alignment \cite{liu2019concept}.
The bootstrapping strategy exploits pattern-concept duality: we can extract a set of concepts from queries following a set of patterns, and we can learn a set of new patterns from a set of queries with extracted concepts. Thus, we can start from a set of seed patterns, and iteratively accumulate more and more patterns and concepts.
The query-title alignment strategy is inspired by the observation that a concept in a query is usually mentioned in the clicked titles associated with the query, yet possibly in a more detailed manner.
Based on this observation, we align a query with its top clicked titles to find a title chunk which fully contains the query tokens in the same order and potentially contains extra tokens within its span. Such a title chunk is selected as a candidate concept.

For events, we split the original unsegmented document titles into subtitles by punctuations and spaces. After that, we only keep the set of subtitles with lengths between $L_l$ (we use 6) and $L_h$ (we use 20). For each remaining subtitle, we score it by counting how many unique non-stop query tokens within it. The subtitles with the same score will be sorted by its click-through rate. Finally, we select the top ranked subtitle as a candidate event phrase.

\textbf{Attention Derivation}.
After extracting a collection of attention nodes (or phrases), we can further derive higher level concepts or topics from them, which automatically become 
their parent nodes
in our Attention Ontology.

On one hand, we derive higher-level concepts by applying Common Suffix Discovery (CSD) to extracted concepts. We perform word segmentation over all concept phrases, and find out the high-frequency suffix words or phrases. If the suffixes forms a noun phrase, we add it as a new concept node. For example, the concept ``animated film (\begin{CJK}{UTF8}{gkai}动画|电影\end{CJK})'' can be derived from ``famous animated film (\begin{CJK}{UTF8}{gkai}著名的|动画|电影\end{CJK})'', ``award-winning animated film (\begin{CJK}{UTF8}{gkai}获奖的|动画|电影\end{CJK})'' and ``Hayao Miyazaki animated film (\begin{CJK}{UTF8}{gkai}宫崎骏|动画|电影\end{CJK})'', as they share the common suffix ``animated film (\begin{CJK}{UTF8}{gkai}动画|电影\end{CJK})''

On the other hand, we drive high-level topics by applying Common Pattern Discovery (CPD) to extracted events. We perform word segmentation, named entity recognition and part-of-speech tagging over the event phrases. Then we find out high-frequency event patterns and recognize the different elements in the events. If the elements (e.g., entities or locations of events) have \textit{isA} relationship with one or multiple common concepts, we replace the different elements by the most fine-grained common concept ancestor in the ontology. For example, we can derive a topic ``Singer will have a concert (\begin{CJK}{UTF8}{gkai}歌手|开|演唱会\end{CJK})'' from ``Jay Chou will have a concert (\begin{CJK}{UTF8}{gkai}周杰伦|开|演唱会\end{CJK})'' and ``Taylor Swift will have a concert (\begin{CJK}{UTF8}{gkai}泰勒斯威夫特|开|演唱会\end{CJK})'', as the two phrases sharing the same pattern ``XXX will have a concert (\begin{CJK}{UTF8}{gkai}XXX|开|演唱会\end{CJK})'', and both ``Jay Chou (\begin{CJK}{UTF8}{gkai}周杰伦\end{CJK})'' and ``Taylor Swift (\begin{CJK}{UTF8}{gkai}泰勒斯威夫特\end{CJK})'' belong to the concept ``Singer (\begin{CJK}{UTF8}{gkai}歌手\end{CJK})''. To ensure that the derived topic phrases are user interests, we filter out phrases that have not been searched by a certain number of users.

\subsection{Linking User Attentions}
\label{sec:connecting}

The previous step produces a large set of nodes representing user attentions (or interests) in different granularities.
Our goal is to construct a taxonomy based on these individual nodes to show their correlations.
With edges between different user attentions, we can reason over it to infer a user's real interest.

In this section, we describe our methods to link attention nodes and construct the complete ontology.
We will construct the \textit{isA} relationships, \textit{involve} relationships, and \textit{correlate} relationships between categories, extracted attention nodes and entities to construct a ontology.
We exploit the following action-driven strategies to link different types of nodes.

\textbf{Edges between Attentions and Categories}.
To identify the \textit{isA} relationship between \textit{attention-category} pairs, we utilize the co-occurrence of them shown in user search click logs.
Given an attention phrase $p$ as the search query, suppose there are $n^p$ clicked documents of search query $p$ in the search click logs,
and among them there are $n^p_g$ documents belong to category $\mathbf{g}$.
We then estimate $\mathbb{P}(g| p)$ by $\mathbb{P}(g | p) = n^{p}_{g} / n^{p}$.
We identify that there is an \textit{isA} relationship between $p$ and $g$ if $\mathbb{P}(g | p) > \delta_g$ (we set $\delta_g= 0.3$).

\textbf{Edges between Attentions}.
To discover \textit{isA} relationships, we utilize the same criteria when we perform attention derivation: we link two concepts by the \textit{isA} relationship if one concept is the suffix of another concept, and we link two topic/event attentions by the \textit{isA} relationship if they share the same pattern and there exists \textit{isA} relationships between their non-overlapping tokens.
Note that if a topic/event doesn't contain an element of another topic/event phrase, it also indicates that they have \textit{isA} relationship. For example, ``Jay Chou will have a concert'' has  \textit{isA} relationship with both ``Singer will have a concert'' and ``have a concert''.
For the \textit{involve} relationship, we connect a concept to a topic if the concept is contained in the topic phrase.

\textbf{Edges between Attentions and Entities}.
We extract: i) \textit{isA} relationships between concepts and entities; ii)  \textit{involve} relationships between topics/events and entities, locations or triggers.

\begin{figure}
\includegraphics[width=0.5\textwidth]{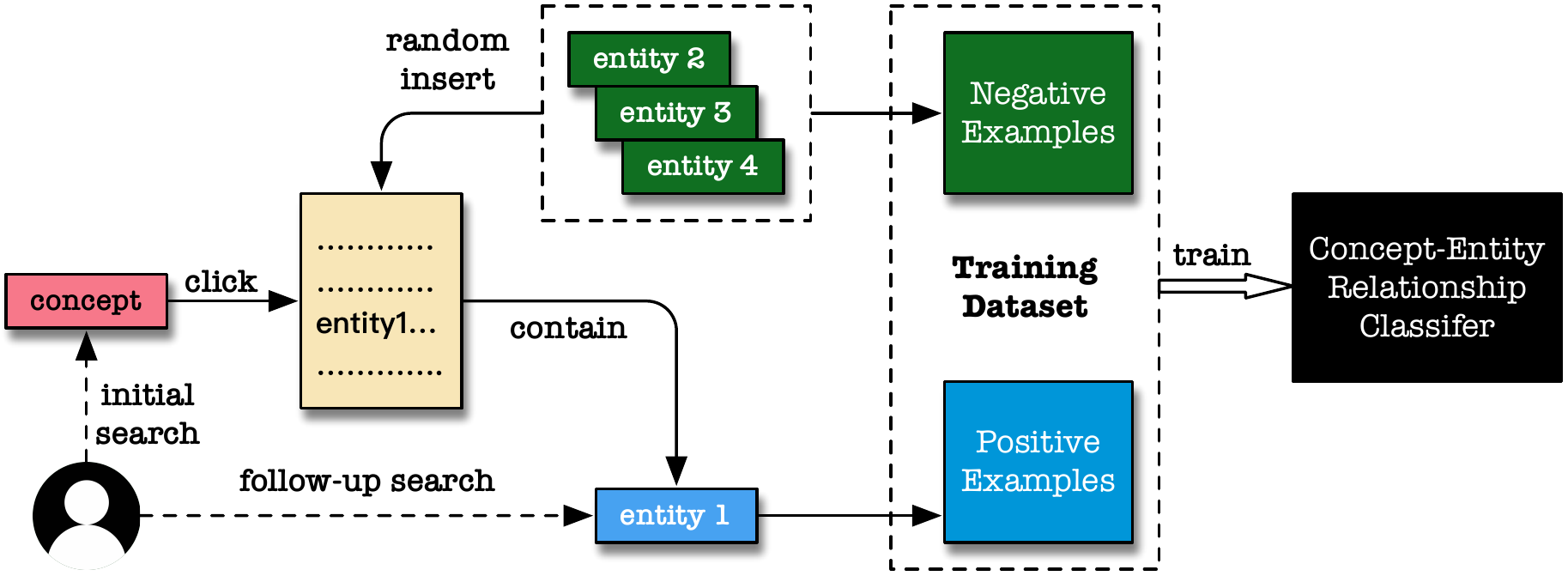} 
\caption{Automatic construction of the training datasets for classifying the \textit{isA} relationship between concepts and entities.}
\Description[Edge Construction]{Automatic construction of the training datasets for classifying the \textit{isA} relationship between concepts and entities.}
\label{fig:concept-entity-edge}
\vspace{-4mm}
\end{figure}

For concepts and entities, using strategies such as co-occurrence will introduce a lot of noises, as co-occurrence doesn't always indicate an entity belongs to a concept.
To solve this issue, we propose to train a concept-entity relationship classifier based on the concept and the entity's context information in the clicked documents.
Labeling a training dataset for this task requires a lot of human efforts.
Instead of manual labeling, we propose a method to automatically construct a training dataset from user search click graphs.
Figure~\ref{fig:concept-entity-edge} shows how we construct such a training dataset. We select the concept-entity pairs from search logs as positive examples if: i) the concept and the entity are two consecutive queries from one user, and ii) the entity is mentioned by a document which a user clicked after issuing the concept as a search query.
Besides, we select entities belonging to the same higher-level concept or category, and insert them into random positions of the document to create negative examples of the dataset.
For the classifier, we can train a classifier such as GBDT based on manual features, or fine-tune a pre-trained language model to incorporate semantic features and infer whether the context indicates a \textit{isA} relationship between the concept-entity pair.

For events/topics and entities, we only recognize the important entities, triggers and locations in the event/topic, and connect them by an \textit{involve} relationship edge.
We first create an initial dataset by extracting all the entities, locations, and trigger words in the events/topic based on a set of predefined trigger words and entities.
Then the dataset is manually revised by workers to remove the unimportant elements.
Based on this dataset, we reuse our GCTSP-Net and train it without ATSP-decoding to perform $4$-class (entity, location, trigger, other) node classification over the query-title interaction graphs of the events/topics. In this way, we can recognize the different elements of an event or topic, and construct \textit{involve} edges between them.

\textbf{Edges between Entities}.
Finally, we construct the \textit{correlate} relationship between entity pairs by the co-occurrence information in user queries and documents.
We utilize high frequency co-occurring entity pairs in queries and documents as positive entity pairs, and perform negative sampling to create negative entity pairs.
After automatically created a dataset from search click logs and web documents, we learn the embedding vectors of entities with Hinge loss, so that the Euclidean distance between two correlated entities will be small.
After learned the embedding vectors of different entities, we classify a pair of entities as correlated if their Euclidean distance is smaller than a threshold.

Note that the same approach for \textit{correlate} relationship discovery can be applied to other type of nodes such as concepts. Currently, we only constructed such relationships between entities.

\section{Applications}
\label{sec:text-undertanding}

In this section, we show how our attention ontology can be applied to a series of NLP tasks to achieve user-centered text understanding.

\textbf{Story Tree Formation}.
The relationships between events and the involved entities, triggers and locations can be utilized to cluster correlated events and form a story tree \cite{liu2017growing}.
A story tree organizes a collection of related events with a tree structure to highlight the evolving nature of a real-world story.
Given an event $p^e$, we retrieve a collection of related events $P^e$ and use them to form a story tree, which allows us to better serve the users by recommending follow-up events from $P^e$ when they have read news articles about $p^e$.

Constructing a story tree from an attention ontology involves four steps: retrieving correlated events, calculating similarity matrix, hierarchical clustering, and tree formation.
First, with the help of the attention ontology, we can retrieve related events set $P^e$ give an event $p^e$. Specifically, the criteria to retrieve ``correlated'' events can be flexible.
For example, we can set a requirement that each event $p_i \in P^e$ shares at least one common child entity with $p^e$, or we can force the triggers of them to be the same.
Second, we can estimate the similarities between each pair of events based on the text similarity of event phrases and the common entities, triggers or locations shared by them.
Specifically, we calculate the similarity between two events by:
\begin{align}
\allowdisplaybreaks
s(p^e_1,\ p^e_2) &= f_m(p^e_1,\ p^e_2) + f_g(p^e_1,\ p^e_2) + f_e(p^e_1,\ p^e_2),\\
f_m(p^e_1,\ p^e_2) &= \text{CosineSimilarity}(\mathbf{v}^{p^e_1},\ \mathbf{v}^{p^e_2}),\\
f_g(p^e_1,\ p^e_2) &= \text{CosineSimilarity}(\mathbf{v}^{g^e_1},\ \mathbf{v}^{g^e_2}),\\
f_e(p^e_1,\ p^e_2) &= \text{TF-IDFSimilarity}(E^{p^e_1},\ E^{p^e_2}),
\end{align}
where $s(p^e_1,\ p^e_2)$ is the measured similarity between the two events, It is given by the sum of three scores: i) $f_m(p^e_1,\ p^e_2)$, which represents the semantic distance between two event phrases. We use the cosine similarity of BERT-based phrase encoding vectors $\mathbf{v}^{p^e_1}$ and $\mathbf{v}^{p^e_2}$ for the two events \cite{devlin2018bert} ; ii) $f_g(p^e_1,\ p^e_2)$, which represents the similarity of the triggers in two events. We calculate the similarity between trigger $g^e_1$ in event $p^e_1$ and $g^e_2$ in $p^e_2$ by the cosine similarity of the  word vectors $\mathbf{v}^{g^e_1}$ and $\mathbf{v}^{g^e_2}$ from \cite{song2018directional}; iii) $f_e(p^e_1,\ p^e_2)$, the TF-IDF similarity between the set of entities $E^{p^e_1}$ of event $p^e_1$ and $E^{p^e_2}$ of $p^e_2$.
After the measurement of the similarities between events, we perform hierarchical clustering to group them into hierarchical clusters.
Finally, we order the events by time, and put the events in the same cluster into the same branch. In this way, the cluster hierarchy is transformed into the branches of a tree.

\textbf{Document Tagging}
We can also utilize the attention phrases to describe the main topics of a document by tagging the correlated attentions to the document, even if the phrase is not explicitly mentioned in the document.
For example, a document talking about films ``Captain America: The First Avenger'', ``Avengers: Endgame'' and ``Iron Man'' can be tagged with the concept ``Marvel Super Hero Movies'' even though the concept may not be contained by it. Similarly, a document talking about ``Theresa May’s Resignation Speech'' can be tagged by topics ``Brexit Negotiation'', while traditional keyword-based methods are not able to reveal such correlations.

To tag concepts to documents, we combine a matching-based approach and a probabilistic inference-based approach based on the key entities in a document.
Suppose $d$ contains a set of key entities $E^d$.
For each entity $e^d \in E^d$, we obtain its parent concepts $P^c$ in the attention ontology as candidate tags. For each candidate concept $p^c \in P^c$, we score the coherence between $d$ and $p^c$ by calculating the TF-IDF similarity between the title of $d$ and the context-enriched representation of $p^c$ (i.e., the topic clicked titles of $p^c$).

When no parent concept can be found by the attention ontology, we identify relevant concepts by utilizing the context information of the entities in $d$.
Denote the probability that concept $p^c$ is related to document $d$ as $\mathbb{P}(p^c|d)$. We estimate it by:
\begin{align}
\allowdisplaybreaks
\mathbb{P}(p^c|d) &= \sum_{i=1}^{|E^d|}\mathbb{P}(p^c|e^d_i) \mathbb{P}(e^d_i | d),\\
\mathbb{P}(p^c|e^d_i) &= \sum_{j = 1}^{|X_{e^d_i}|} \mathbb{P}(p^c | x_j) \mathbb{P}(x_j | e^d_i),\\
\mathbb{P}(p^c | x_j) &= 
\begin{cases}\frac{1}{|P^c_{x_j}|} \ \   \text{ if } x_j \text{ is a substring of } p^c,\\
0  \ \ \ \ \ \ \ \ \text{otherwise.}
\end{cases}
\end{align}
where $E^d$ is the key entities of $d$, $\mathbb{P}(e^d_i | d)$ is the document frequency of entity $e^d_i \in E^d$.
$\mathbb{P}(p^c|e^d_i)$ estimates the probability of concept $p^c$ given $e^d_i$, which is inferred from the context words of $e^d_i$.
$\mathbb{P}(x_j | e^d_i)$ is the co-occurrence probability of context word $x_j$ with $e^d_i$. We consider two words as co-occurred if they are contained in the same sentence. $X_{e^d_i}$ are the set of contextual words of $e^d_i$ in $d$. $P^c_{x_j}$ is the set of concepts containing $x_j$ as a substring.

To tag events or topics to a document, we combine longest common subsequence based (LCS-based) textural matching with Duet-based semantic matching \cite{mitra2017learning}.
For LCS-based matching, we concatenate a document title with the first sentence in content, and calculate the length of longest common subsequence between a topic/event phrase and the concatenated string.
For Duet-based matching, we utilize the Duet neural network \cite{mitra2017learning} to classify whether the phrase matches with the concatenated string.
If the length of the longest common subsequence is above a threshold and the classification result is positive, we tag the phrase to the document.

\textbf{Query Understanding}.
A user used to search about an entity may be interested in a broader class of similar entities. However, the user may not be even aware of the entities similar to the query.
With the help of our ontology, we can better understand users' implicit intention and perform query conceptualization or recommendation to improve the user experience in search engines.
Specifically, we analyze whether a query $q$ contains a concept $p^c$ or an entity $e$.
If a query conveys a concept $p^c$, we can rewrite it by concatenating $q$ with each of the entities $e_i$ that have \textit{isA} relationship with $p^c$.
In this way, we rewrite the query to the format of ``$q$ $e_i$''.
If a query conveys an entity $e$, we can perform query recommendation by recommend users the entities $e_i$ that have \textit{correlate} relationship with $e$ in the ontology.

\section{Evaluation}
\label{sec:eval}

Our proposed GIANT system is the core ontology system in multiple applications in Tencent, i.e., Tencent QQ Browser, Mobile QQ and WeChat, and is serving more than a billion daily active users all around the world.
It is implemented by Python 2.7 and C++.
Each component of our system works as a service and is deployed with Tars\footnote{https://github.com/TarsCloud/Tars}, a high-performance remote procedure call (RPC) framework based on name service and Tars protocol.
The attention mining and linking services are deployed on 10 dockers, with each configured with four processor Intel(R) Xeon(R) Gold 6133 CPU @ 2.50GHz and 6GB memory. Applications such as document tagging are running on 48 dockers with the same configuration. MySQL database is used for data storage.
We construct the attention ontology from large-scale real-world daily user search click logs. While our system is deployed on Chinese datasets, the techniques proposed in our work can be easily adapted to other languages.


\subsection{Evaluation of the Attention Ontology}

\begin{table}[tb]
\small
  \begin{tabular}{l|ccccc}
    \toprule
     & Category & Concept & Topic & Event & Entity\\
    \midrule
    Quantity & $1,206$ & $460,652$ & $12,679$ & $86,253$ & $1,980,841$\\
    Grow / day & - & $11,000$ & - & $120$ & -\\
    \bottomrule
  \end{tabular}
  \caption{Nodes in the attention ontology.}
  \label{tab:node-statistics}
  \vspace{-4mm}
\end{table}

\begin{table}[tb]
\small
  \begin{tabular}{l|ccccc}
    \toprule
     & \textit{isA} & \textit{correlate} & \textit{involve} \\
    \midrule
    Quantity & $490,741$ & $1,080,344$ & $160,485$ \\
    Accuracy & $95\%+$ & $95\%+$ & $99\%+$\\
    \bottomrule
  \end{tabular}
  \caption{Edges in the attention ontology.}
  \label{tab:edge-statistics}
  \vspace{-5mm}
\end{table}

Table~\ref{tab:node-statistics} shows the statistics of different nodes in the attention ontology. We extract attention phrases from daily user search click logs. Therefore, the scale of our ontology keeps growing. Currently, our ontology contains $1,206$ predefined categories, $460,652$ concepts, $12,679$ topics, $86,253$ events and $1,980,841$ entities. We are able to extract around $27,000$ concepts and $400$ events every day, and around $11,000$ concepts and $120$ events are new. For online concept and event tagging, our system processes $350$ documents per second.

Table~\ref{tab:edge-statistics} shows the statistics and accuracies of different types of edges (relationships) in the attention ontology. Currently, we have constructed more than $490K$ \textit{isA} relationships, $1,080K$ \textit{correlate} relationships, and $160K$ \textit{involve} relationships between different nodes. The human evaluation performed by three managers in Tencent shows the accuracies of the three relationship types are above $95\%$, $95\%$ and $99\%$, respectively.

\begin{table}[tb]
\small
  \begin{tabularx}{\columnwidth}{>{\hsize=0.16\columnwidth}X|>{\hsize=0.26\columnwidth}X|X}
    \toprule
    Categories & Concepts & Instances \\
    \midrule
    Sports  & Famous long-distance runner & Dennis Kipruto Kimetto, Kenenisa Bekele\\
    \hline
    Stars   & Actors who committed suicide  & Robin Williams, Zhang Guorong, David Strickland \\
    \hline
   Drama series  & American crime drama series  & American Crime Story, Breaking Bad, Criminal Minds \\
    \hline
    Fiction & Detective fiction & Adventure of Sherlock Holmes, The Maltese Falcon\\
    \bottomrule
  \end{tabularx}
  \caption{Showcases of concepts and the related categories and entities.}
  \label{tab:show-case}
  \vspace{-3mm}
\end{table}

\begin{table}[ht]
\begin{center}
\centering
\small
\begin{tabular}{p{1.2cm}|p{1.5cm}|p{3.2cm}|p{1.5cm}}    
    \toprule
    Categories & Topics & Events & Entities \\
    \midrule
    Music & Singers win music awards & Jay Chou won the Golden Melody Awards in 2002, Taylor Swift  won the 2016 Grammy Awards & Jay Chou, Taylor Swift \\
    \hline
    Cellphone & cellphone launch events & Apple news conferences 2018 mid-season, Samsung Galaxy S9 officially released & Apple, iPhone, Samsung, Galaxy S9 \\
   \hline
   Esports & League of Legends season 8 & LOL S8 finals announcement, IG wins the S8 final, IG's reward for winning the S8 final revealed & League of Legends, IG team, finals \\
    \bottomrule
\end{tabular}
\caption{Showcases of events and the related categories, topics and involved entities.}
\label{tab:show-case-events}
\end{center}
\vspace{-3mm}
\end{table}

To give intuition into what kind of attention phrases can be derived from search click graphs, Table~\ref{tab:show-case} and Table~\ref{tab:show-case-events} show a few typical examples of concepts and events (translated from Chinese), and some topics, categories, and entities that share \textit{isA} relationship with them.
By fully exploiting the information of user actions contained in search click graphs, we transform user actions to user attentions, and extract concepts such as ``\begin{CJK}{UTF8}{gkai}Actors who committed suicide (自杀的演员)\end{CJK}''.
Besides, we can also extract events or higher level topics of users' interests, such as ``Taylor Swift and Katy Perry'' if a user often inputs related queries. Based on the connections between entities, concepts, events and topics, we can also infer what a user really cares and improve the performance of recommendation.

\subsection{Evaluation of the GCTSP-Net}

We evaluate our GCTSP-Net on multiple tasks by comparing it to a variety of baseline approaches.

\textbf{Datasets}.
To the best of our knowledge, there is no publicly available dataset suitable for the task of heterogeneous phrase mining  from user queries and search click graphs.
To construct the user attention ontology, we create two large-scale datasets for concept mining and event mining using the approaches described in Sec.~\ref{sec:attention}: the Concept Mining Dataset (CMD) and the Event Mining Dataset (EMD).
These two datasets contain $10,000$ examples and $10,668$ examples respectively. Each example is composed of a set of correlated queries and top clicked document titles from real-world query logs, together with a manually labeled gold phrase (concept or event).
For the event mining dataset, it further contains triggers, key entities and locations of each event. We use the earliest article publication time as the time of each event example.
The datasets are labeled by $3$ professional product managers in Tencent and $3$ graduate students.
For each dataset, we utilize $80\%$ as training set, $10\%$ as development set, and $10\%$ as test set.
The datasets will be published for research purposes \footnote{https://github.com/BangLiu/GIANT}.


\textbf{Methodology and Models for Comparison}. We compare our GCTSP-Net with the following baseline methods on the concept mining task:
\begin{itemize}
    \item \textbf{TextRank}. A classical graph-based keyword extraction model \cite{mihalcea2004textrank}.\footnote{https://github.com/letiantian/TextRank4ZH}
    \item \textbf{AutoPhrase}. A state-of-the-art phrase mining algorithm that extracts quality phrases based on POS-guided segmentation and knowledge base \cite{shang2018automated}.\footnote{https://github.com/shangjingbo1226/AutoPhrase}
  \item \textbf{Match}. Extract concepts from queries and titles by matching using patterns from bootstrapping  \cite{liu2019concept}.
  \item \textbf{Align}. Extract concepts by the query-title alignment strategy described in Sec.~\ref{sec:mining}.
  \item \textbf{MatchAlign}. Extract concepts by both pattern matching and query-title alignment strategy.
  \item \textbf{LSTM-CRF-Q}. Apply LSTM-CRF \cite{huang2015bidirectional} to input query.
  \item \textbf{LSTM-CRF-T}. Apply LSTM-CRF \cite{huang2015bidirectional} to titles.
\end{itemize}
For the TextRank and AutoPhrase algorithm, we extract the top 5 keywords or phrases from queries and titles, and concatenate them in the same order with the query/title to get the extracted phrase.
For MatchAlign, we select the most frequent result if multiple phrases are extracted.
For LSTM-CRF-Q/LSTM-CRF-T, it consists of a 200-dimensional word embedding layer initialized with the word vectors proposed in \cite{song2018directional}, a BiLSTM layer with hidden size 25 for each direction, and a Conditional Random Field (CRF) layer which predicts whether each word belongs to the output phrase by Beginning-Inside–Outside (BIO) tags.

For event mining task, we compare with TextRank and LSTM-CRF. In addition, we also compare with:
\begin{itemize}
  \item \textbf{TextSummary \cite{mihalcea2004textrank}}. An encoder-decoder model with attention mechanism for text summarization.\footnote{https://github.com/dongjun-Lee/text-summarization-tensorflow}
  \item \textbf{CoverRank}. Rank queries and subtitles by counting the covered nonstop query words, as described in \ref{sec:mining}.
\end{itemize}
For TextRank, we use the top $2$ queries and top $2$ selected subtitles given by CoverRank, and perform re-ranking.
For TextSummary, we use the 200-dimensional word embeddings in \cite{song2018directional}, two-layer BiLSTM (256 hidden size for each direction) as encoder, and one layer LSTM with 512 hidden size and attention mechanism as decoder (beam size for decoding is 10).
We feed the concatenation of queries and titles into TextSummary to generate the output.
For LSTM-CRF, the LSTM layer in it is configured similarly to the encoder of TextSummary.
We feed each title individually into it to get different outputs, filter the outputs by length (number of characters between 6 and 20), and finally select the phrase which belongs to the top clicked title.

For event key elements recognition (key entities, trigger, location), it is a 4-class classification task over each word. We compare our model with LSTM and LSTM-CRF. The difference between LSTM and LSTM-CRF is that LSTM replaces the CRF layer in LSTM-CRF with a softmax layer.

For each baseline, we individually tune the hyper-parameters in order to achieve its best performance. As to our approach (GCTSP-Net), we stack $5$-layer R-GCN with hidden size $32$ and number of bases $B=5$ in basis decomposition for graph encoding and node classification. We will open-source our code together with the datasets for research purposes.

\textbf{Metrics}. We use Exact Match (EM), F1 and coverage rate (COV) to evaluate the performance of phrase mining tasks. The exact match score is 1 if the prediction is exactly the same as ground-truth or 0 otherwise.
F1 measures the portion of overlap tokens between the predicted phrase and the ground-truth phrase \cite{rajpurkar2016squad}.
The coverage rate measures the percentage of non-empty predictions of each approach.
For the event key elements recognition task, we evaluate by the F1-macro, F1-micro, and F1-weighted metrics.

\begin{table}[tb]
\small
  \begin{tabular}{l|lll}
    \toprule
    Method & EM & F1 & COV\\
    \midrule
    TextRank & $0.1941$ & $0.7356$ & $1$ \\
    AutoPhrase  & $0.0725$ & $0.4839$ & $0.9353$ \\
    Match  & $0.1494$ & $0.3054$ & $0.3639$ \\
    Align  & $0.7016$ & $0.8895$ & $0.9611$ \\
    MatchAlign  & $0.6462$ & $0.8814$ & $0.9700$\\
    Q-LSTM-CRF  & $0.7171$ & $0.8828$ & $0.9731$ \\
    T-LSTM-CRF  & $0.3106$ & $0.6333$ & $0.9062$ \\
    \midrule
    GCTSP-Net  & $\mathbf{0.783}$ & $\mathbf{0.9576}$ & $\mathbf{1}$ \\
    \bottomrule
  \end{tabular}
  \caption{Compare concept mining approaches.}
  \label{tab:concept-mining}
  \vspace{-3mm}
\end{table}

\begin{table}[tb]
\small
  \begin{tabular}{l|lll}
    \toprule
    Method & EM & F1 & COV \\
    \midrule
    TextRank & $0.3968$ & $0.8102$ & $1$ \\
    CoverRank  & $0.4663$ & $0.8169$ & $1$\\
    TextSummary  & $0.0047$ & $0.1064$ & $1$ \\
    LSTM-CRF  & $0.4597$ & $0.8469$ & $1$ \\
    \midrule
    GCTSP-Net  & $\mathbf{0.5164}$ & $\mathbf{0.8562}$ & $0.9972$ \\
    \bottomrule
  \end{tabular}
  \caption{Compare event mining approaches.}
  \label{tab:event-mining}
  \vspace{-5mm}
\end{table}

\begin{table}[tb]
\small
  \begin{tabular}{l|lll}
    \toprule
    Method & F1-macro & F1-micro & F1-weighted \\
    \midrule
    LSTM & $0.2108$ & $0.5532$ & $0.6563$ \\
    LSTM-CRF  & $0.2610$ & $0.6468$ & $0.7238$\\
    \midrule
    GCTSP-Net  & $\mathbf{0.6291}$ & $\mathbf{0.9438}$ & $\mathbf{0.9331}$ \\
    \bottomrule
  \end{tabular}
  \caption{Compare event key elements recognition approaches.}
  \label{tab:element-mining}
  \vspace{-5mm}
\end{table}

\textbf{Evaluation results and analysis.} Table \ref{tab:concept-mining}, Table \ref{tab:event-mining}, and Table \ref{tab:element-mining} compare our model with different baselines on the CMD and EMD datasets for concept mining, event mining, and event key elements recognition.
We can see that our unified model for different tasks significantly outperforms all baseline approaches.
The outstanding performance can be attribute to:
first, our query-title interaction graph efficiently encodes the information of word adjacency, word features, dependencies, query-title overlapping and so on in a structured manner, which are critical for both attention phrase mining tasks and event key elements recognition tasks.
Second, the multi-layer R-GCN encoder in our model can learn from both the node features and the multi-resolution structural patterns from query-title interaction graph.
Therefore, combining query-title interaction graph with R-GCN encoder, we can achieve great performance in different node classification tasks. Furthermore, the unsupervised ATSP-decoding sorts the extracted tokens to form an ordered phrase efficiently.
In contrast, heuristic-based approaches and LSTM-based approaches are not robust to the noises in dataset, and cannot capture the structural information in queries and titles.
In addition, existing keyphrase extraction methods such as TextRank \cite{mihalcea2004textrank} and AutoPhrase \cite{shang2018automated} are better suited for extracting key words or phrases from long documents, lacking the ability to give satisfactory performance in our tasks.


\subsection{Applications: Story Tree Formation and Document Tagging}

\begin{figure}
\includegraphics[width=0.5\textwidth]{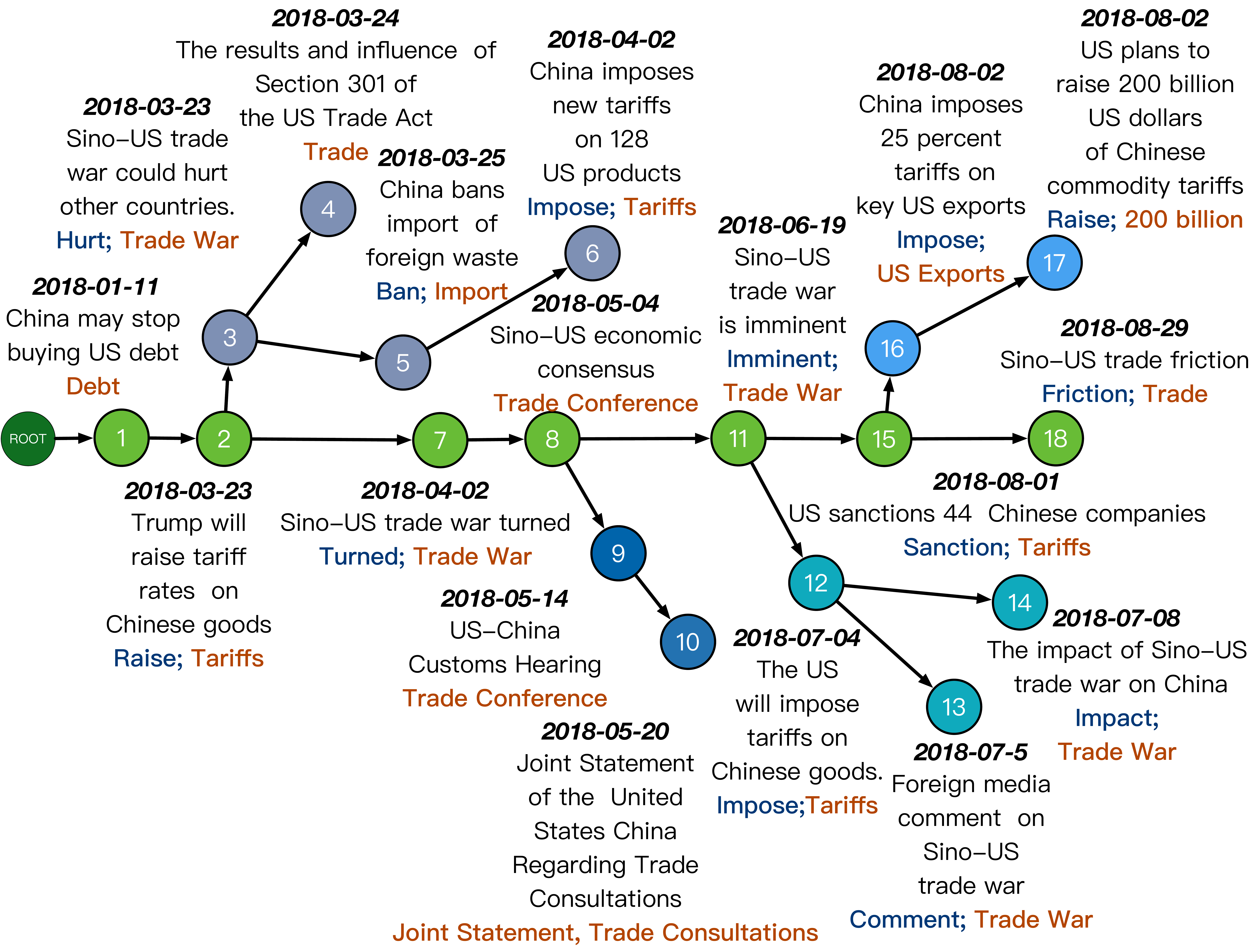} 
\caption{An example to show the constructed story tree given by our approach.}
\Description[Story Tree]{An example to show the constructed story tree given by our approach.}
\label{fig:story}
\vspace{-5mm}
\end{figure}

\textbf{Story Tree Formation}.
We apply the story tree formation algorithm described in Sec.~\ref{sec:text-undertanding} to real-world events to test its performance. Figure~\ref{fig:story} gives an example to illustrate what we can obtain through our approach.
In the example, each node is an event, together with the documents that can be tagged by this event.
We can see that our method can successfully cluster  events related to ``China-US Trade'', ordering the events by the published time of the articles, and show the evolving structure of coherent events.
For example, the branch consists of events $3\sim6$ are mainly about ``Sino-US trade war is emerging'', the branch $8\sim10$ are resolving around ``US agrees limited trade deal with China'', and $11\sim14$ are about ``Impact of Sino-US trade war felt in both countries''.
By clustering and organizing events and related documents in such a tree structure, we can track the development of different stories (clusters of coherent events), reduce information redundancy, and improve document recommendation by recommending users the follow-up events they are interested in.

\textbf{Document Tagging}. For document tagging, our system currently processes around $1,525,682$ documents per day, where about $35\%$ of them can be tagged with at least one concept, and $4\%$ can be tagged with an event.
We perform human evaluation by sampling $500$ documents for each major category (``game'', ``technology'', ``car'', and ``entertainment'').
The result shows that the precision of concept tagging for documents is $88\%$ for ``game'' category, $91\%$ for ``technology'', $90\%$ for ``car'', and $87\%$ for ``entertainment''. The overall precision for documents of all categories is $88\%$.
As to event tagging on documents, the overall precision is $96\%$.

\subsection{Online Recommendation Performance}

We evaluate the effect of our attention ontology on recommendations by analyzing its performance in the news feeds stream of Tencent QQ Browser which has more than $110$ million daily active users.
In the application, both users and articles are tagged with categories, topics, concepts, events or entities from the attention ontology, as shown in Figure~\ref{fig:A3}.
The application recommends news articles to users based on a variety of strategies, such as content-based recommendation, collaborative filtering and so on.
For the content-based recommendation, it matches users with articles through the common tags they share. 

\begin{figure}
\includegraphics[width=0.44\textwidth]{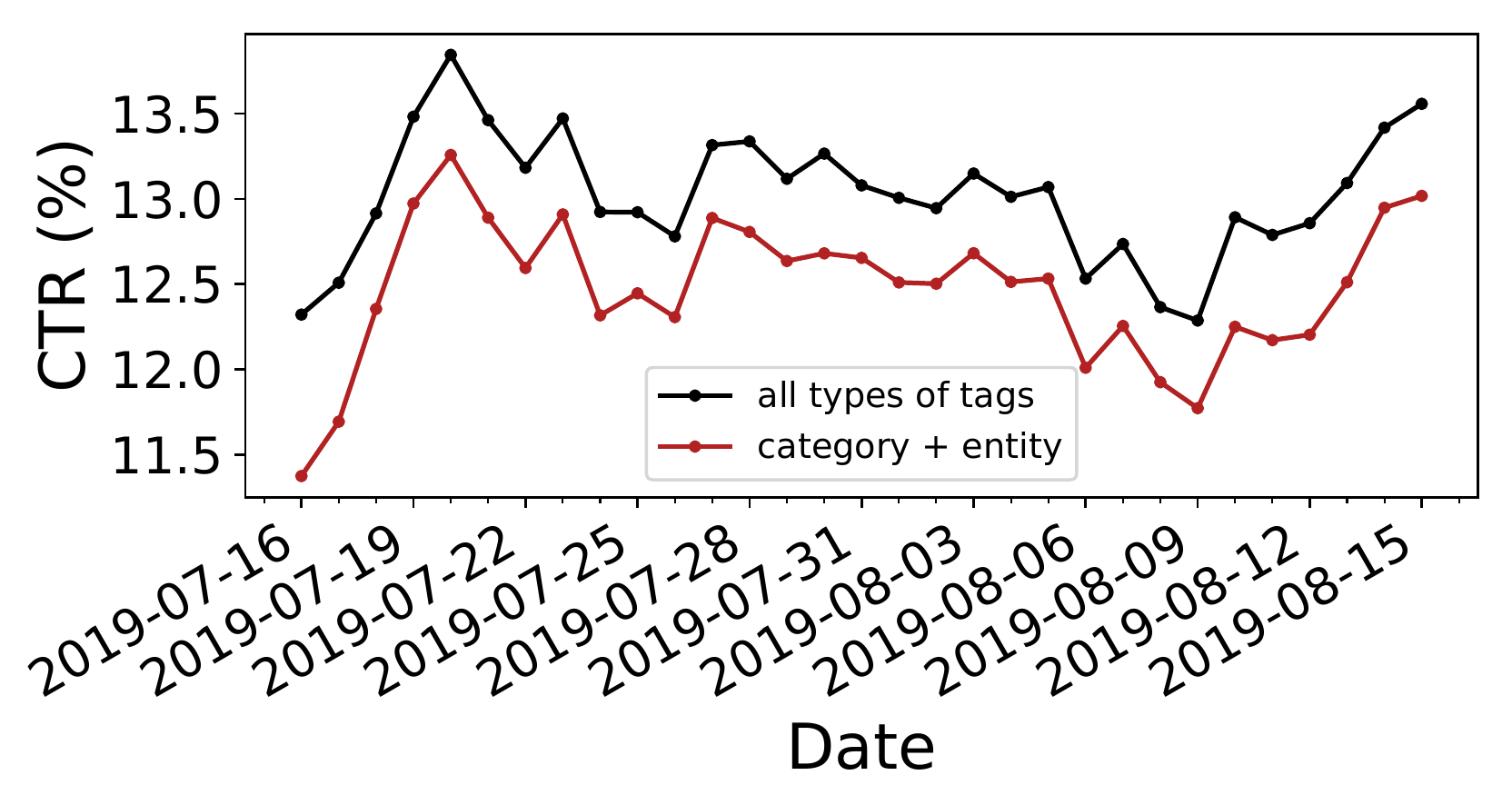} 
\caption{The click-through rates with/without extracted tags.}
\Description[CTR2]{The click-through rates with/without attention.}
\label{fig:ctr1}
\vspace{-5mm}
\end{figure}

\begin{figure}
\includegraphics[width=0.44\textwidth]{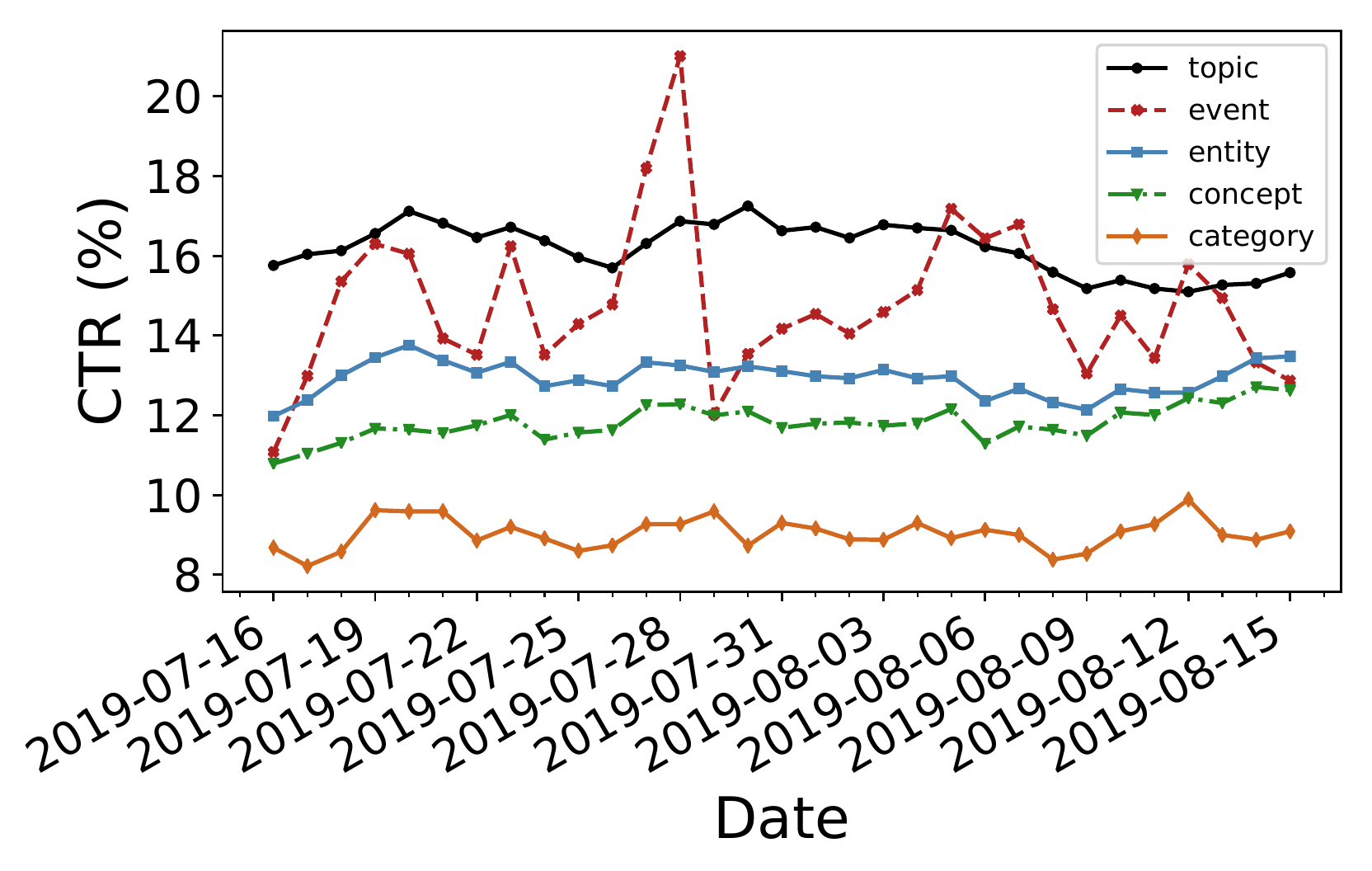} 
\caption{The click-through rates of different tags.}
\Description[CTR1]{The click-through rates of different tags.}
\label{fig:ctr2}
\vspace{-5mm}
\end{figure}

We analyze the Click-Through Rate (CTR) given by different tags from July 16, 2019 to August 15, 2019 to evaluate their effects.
Click-through rate is the ratio of the number of users who click on a recommended link to the number of total users who received the recommendation.
Figure~\ref{fig:ctr1} compares the CTR when we recommend documents to users with different strategies.
Traditional recommender systems only utilize the category tags and entity tags.
We can see that including more types of attention tags (topics, events, or concepts) in recommendation can constantly help improving the CTR on different days: the average CTR improved from 12.47\% to 13.02\%.
The reason is that the extracted concepts, events or topics can depict user interests with suitable granularity and help to solve the inaccurate recommendation and monotonous recommendation problems.

We further analyze the effects of each type of tags in recommendation.
Figure~\ref{fig:ctr2} shows the CTR of the recommendations given by different types of tags.
The average CTR for topic, event, entity, concept and category are 16.18\%, 14.78\%, 12.93\%, 11.82\%, and 9.04\%, respectively.
The CTR of both topic tags and event tags are much higher than that of category and entities, which shows the effectiveness of our constructed ontology.
For events, the events happening on each days dramatically different with each other, and they are not always attractive to users. Therefore, the CTR of event tags is less stable than the topic tags.
For concept tags, they are generalization of fine-grained entities and has \textit{isA} relationship with them. As there may have noises when we perform user interest inference using the relationships between entities and concepts, the CTR of concepts are slightly lower than entities.
However, compared to entities, our experience show that concepts can significantly increase the diversity of recommendation and are helpful in solving the problem of monotonous recommendation.

\section{Related Work}
\label{sec:related}
Our work is mainly related to the following research lines.

\textbf{Taxonomy and Knowledge Base Construction}.
Most existing taxonomy or knowledge bases, such as Probase \cite{wu2012probase}, DBPedia \cite{lehmann2015dbpedia}, YAGO \cite {suchanek2007yago}, extract general concepts about the world and construct graphs or taxonomies based on Wikipedia or formal documents. In contrast, our work utilizes search click graphs to construct an ontology for describing user interests or attentions.
Our prior work \cite{liu2019concept} constructs a three-layered  taxonomy from search logs. Compared to it, our GIANT system constructs an attention ontology with more types of nodes and relationships based a novel algorithm for heterogeneous phrase mining.
There are also works that construct a taxonomy from keywords \cite{liu2012automatic} or queries \cite{baeza2007extracting}. Biperpedia \cite{gupta2014biperpedia} extracts class-attribute pairs from query stream to expand a knowledge graph. \cite{pasca2008weakly} extracts classes of instances with attributes and class labels from web documents and query logs.

\textbf{Concept Mining}.
Existing approaches on concept mining are closely related to research works on named entity recognition \cite{nadeau2007survey,ritter2011named,lample2016neural}, term recognition \cite{frantzi2000automatic,park2002automatic,zhang2008comparative}, keyphrase extraction \cite{witten2005kea,el2014scalable} or quality phrase mining \cite{liu2015mining,shang2018automated,liu2019concept}.
Traditional algorithms utilize pre-defined part-of-speech (POS) templates and dependency parsing to identify noun phrases as term candidates \cite{koo2008simple,shang2018automated}.
Supervised noun phrase chunking techniques \cite{chen1994extracting,punyakanok2001use} automatically learn rules for identifying noun phrase boundaries.
There are also methods that utilize resources such as knowledge graph to further enhance the precision \cite{witten2006thesaurus,ren2017cotype}.
Data-driven approaches make use of frequency statistics in the corpus to generate candidate terms and evaluate their quality \cite{parameswaran2010towards,el2014scalable,liu2015mining}.
Phrase quality-based approaches exploit statistical features to measure phrase quality, and learn a quality scoring function by using knowledge base entity names as training labels \cite{liu2015mining,shang2018automated}.
Neural network-based approaches consider the problem as sequence tagging and train complex deep neural models based on CNN or LSTM-CRF \cite{huang2015bidirectional}.

\textbf{Event Extraction}.
Existing research works on event extraction aim to identify different types of event triggers and their arguments from unstructured text data.
They combine supervised or semi-supervised learning with features derived from training data to classify event types, triggers and arguments \cite {ji2008refining,chen2017automatically,liu2016leveraging,nguyen2016joint,huang2012bootstrapped}.
However, these approaches cannot be applied to new types of events without additional annotation effort.
The ACE2005 corpus \cite{grishman2005nyu} includes  event annotations for 33 types of events. However, such small hand-labeled data is hard to train a model to extract maybe thousands of event types in real-world scenarios.
There are also works using neural networks such as RNNs \cite{nguyen2016joint,sha2018jointly}, CNNs \cite{chen2015event,nguyen2016modeling} or GCNs \cite{liu2018jointly} to extract events from text.
Open domain event extraction \cite{valenzuela2015domain,ritter2012open} extracts news-worthy clusters of words, segments and frames from social media data such as Twitter \cite{atefeh2015survey}, usually under unsupervised or semi-supervised settings and exploits information redundancy.

\textbf{Relation Extraction}.
A comprehensive introduction about relation extraction can be found in \cite{pawar2017relation}.
Most existing techniques for relation extraction can be classified into the following classes.
First, supervised learning techniques, such as features-based \cite{guodong2005exploring} and kernel based \cite{culotta2004dependency} approaches, require entity pairs that labeled with one of the pre-defined relation types as the training dataset.
Second, semi-supervised approaches, including bootstrapping \cite{brin1998extracting}, active learning \cite{liu2016effective,settles2009active} and label propagation \cite{chen2006relation}, exploit the unlabeled data to reduce the manual efforts of creating large-scale labeled dataset.
Third, unsupervised methods \cite{yan2009unsupervised} utilize techniques such as clustering and named entity recognition to discover relationships between entities.
Fourth, Open Information Extraction \cite{fader2011identifying} construct comprehensive systems to automatically discover possible relations of interest using text corpus.
Last, distant supervision based techniques  leverage pre-existing structured or semi-structured data or knowledge to guide the extraction process \cite{zeng2015distant,smirnova2018relation}.


\section{Conclusion}
\label{sec:conclude}

In this paper, we describe our design and implementation of GIANT, a system that proposed to construct a web-scale user attention ontology from large amount of query logs and search click graphs for various applications.
It consists of around two millions of  heterogeneous nodes with three types of relationships between them, and keeps growing with newly retrieved nodes and identified relationships every day.
To construct the ontology, we propose the query-title interaction graph to represent the correlations (such as adjacency or syntactical dependency) between the tokens in correlated queries and document titles.
Furthermore, we propose the GCTSP-Net to extract multi-type phrases from the query-title interaction graph, as well as recognize the key entities, triggers or locations in events.
After constructing the attention ontology by our models, we apply it to different applications, including document tagging, story tree formation, as well as recommendation.
We run extensive experiments and analysis to evaluate the quality of the constructed ontology and the performance of our new algorithms. Results show that our approach outperforms a variety of baseline methods on three tasks. In addition, our attention ontology significantly improves the CTR of news feeds recommendation in a real-world application.

\bibliographystyle{ACM-Reference-Format}
\balance
\bibliography{main}

\clearpage

\end{document}